%% file: main_icme.tex
\documentclass[conference]{IEEEtran}
\IEEEoverridecommandlockouts

\usepackage{cite}
\usepackage{amsmath,amssymb,amsfonts}
\usepackage{algorithmic}
\usepackage{graphicx}
\usepackage{textcomp}
\usepackage{xcolor}

\usepackage{times} 
\usepackage{epsfig}
\usepackage{tabularx}
\usepackage{adjustbox}
\usepackage{colortbl}
\usepackage{booktabs}
\usepackage{placeins}
\usepackage{array}
\usepackage{makecell}
\usepackage[stretch=10,shrink=10,tracking=true]{microtype}
\usepackage{xurl}
\usepackage{breakcites}
\usepackage{seqsplit}
\usepackage{algorithm} 
\usepackage{enumerate}
\usepackage{multirow}
\usepackage{multicol}
\usepackage{arydshln}
\usepackage{soul}
\usepackage{subfig} % IEEEtran 有时推荐 subfigure 或 subcaption，但在 conference 模式下通常兼容
\usepackage{caption}

\def\BibTeX{{\rm B\kern-.05em{\sc i\kern-.025em b}\kern-.08em
    T\kern-.1667em\lower.7ex\hbox{E}\kern-.125emX}}

\input{defs}

\newtheorem{theorem}{\textbf{Theorem}}

\begin{document}

% --------------------------------------------------------------------------
% 3. 标题与作者 (Title & Authors)
% --------------------------------------------------------------------------
%\title{{\huge Graph Algorithm Unrolling with Douglas-Rachford Iterations for Image Interpolation with Informed Initialization}}
\title{{Unrolling Graph-based Douglas-Rachford Algorithm for Image Interpolation with Informed Initialization}}

% ICME 双盲审稿 (Double-blind review)，提交时使用匿名作者
%\author{Anonymous ICME submission}

% 最终版 (Camera-Ready)
\author{

    Xue Zhang$^{1,2,3}$, Bingshuo Hu$^{1,2,3}$, Gene Cheung$^{4,*}$ \\
    {\small $^1$Shandong Provincial Key Laboratory of Smart Mine Information Technology, China} \\
    {\small $^2$Qingdao Key Laboratory of Intelligent Sensing and Virtual Emergency Simulation, China} \\
    {\small $^3$College of Computer Science and Engineering, Shandong University of Science and Technology, China} \\
    {\small $^4$Dept of EECS, York University, Canada} 
    % 底部注脚部分
    \thanks{* Corresponding author: Gene Cheung (genec@yorku.ca).}
    \thanks{The work of Xue Zhang was supported by the National Natural Science Foundation of China under Grant 62302278, the Natural Science Foundation of Shandong Province under Grant ZR2023QF014, the Higher Education Institutions Youth Innovation and Science \& Technology Support Program of Shandong Province under Grant 2024KJH066, and the Young Talent of Lifting engineering for Science and Technology in Shandong under Grant SDAST2024QTA055. The work of G. Cheung was supported in part by the Natural Sciences and Engineering Research Council of Canada (NSERC) RGPIN-2025-06252. }
}
\maketitle

% --------------------------------------------------------------------------
% 4. 摘要与关键词 (Abstract & Keywords)
% --------------------------------------------------------------------------
\begin{abstract}
Conventional deep neural nets (DNNs) initialize network parameters at random and then optimize each one via stochastic gradient descent (SGD), resulting in substantial risk of poor-performing local minima.
Focusing on image interpolation and leveraging a recent theorem that maps a (pseudo-)linear interpolator $\bTheta$ to a directed graph filter that is a solution to a corresponding MAP problem with a graph shift variation (GSV) prior, we first initialize a directed graph adjacency matrix $\A$ given a known interpolator $\bTheta$, establishing a baseline performance.
Then, towards further gain, we learn perturbation matrices $\P$ and $\P^{(2)}$ from data to augment $\A$, whose restoration effects are implemented progressively via Douglas-Rachford (DR) iterations, which we unroll into a lightweight and interpretable neural net.
Experiments on different image interpolation scenarios demonstrate state-of-the-art performance, while drastically reducing network parameters and inference complexity.
\end{abstract}

% 注意：IEEEtran 使用 IEEEkeywords 环境

% --------------------------------------------------------------------------
% 5. 正文内容 (Main Content)
% --------------------------------------------------------------------------
% 保持你的模块化结构，直接调用 ICASSP26 文件夹下的内容
% 只有当子文件内部包含 \ninept 或 spconf 特有命令时才需要修改子文件

\section{Introduction}
\label{sec:intro}

\input{intro}

\section{Preliminaries}
\label{sec:prelim}
\input{prelim}

\section{Problem Formulation}
\label{sec:formulation}
\input{formulate}

\section{Algorithm Unrolling}
\label{sec:unrolling}
\input{unrolling}

\section{Experiments}
\label{sec:results}

\input{results}

\section{Conclusion}
\label{sec:conclude}
\input{conclude}

\appendix
\input{append} 

% \section{Rebuttal}
% \label{sec:rebuttal}
% \input{rebuttal}

\begin{small}

\bibliographystyle{IEEEbib}
\bibliography{ref2} % 确保 ref2.bib 文件在同一目录下

\end{small}

\end{document}

%% file: defs.tex
\def\0{{\mathbf 0}}
\def\1{{\mathbf 1}}

\def\f{{\mathbf f}}

\def\u{{\mathbf u}}
\def\v{{\mathbf v}}

\def\x{{\mathbf x}}
\def\y{{\mathbf y}}
\def\z{{\mathbf z}}

\def\A{{\mathbf A}}
\def\B{{\mathbf B}}
\def\C{{\mathbf C}}
\def\D{{\mathbf D}}

\def\G{{\mathbf G}}

\def\H{{\mathbf H}}
\def\I{{\mathbf I}}

\def\M{{\mathbf M}}

\def\P{{\mathbf P}}
\def\Q{{\mathbf Q}}
\def\R{{\mathbf R}}

\def\W{{\mathbf W}}

\def\ie{{\textit{i.e.}}}
\def\eg{{\textit{e.g.}}}

\def\cG{{\mathcal G}}

\def\cO{{\mathcal O}}

\def\bTheta{{\boldsymbol \Theta}}

%% file: intro.tex
Image interpolation---a well-studied image restoration task---estimates missing pixel values at targeted 2D grid locations given observed neighboring image pixels.
While early methods include model-based ones such as linear and bicubic interpolations\,\cite{shewchuk94q}, recent methods are dominated by deep-learning-based models such as SwinIR\,\cite{Liang_2021_ICCV} and Restormer\,\cite{zamir2022restormer}, driven by powerful off-the-shelf neural architectures, such as \textit{convolutional neural nets} (CNNs)
%\,\cite{2012ImageNet} 
and transformers\,\cite{vaswani17attention}. 
However, these models require huge numbers of network parameters, leading to large training, storage, and inference costs. 
Moreover, initialization of network parameters is often done randomly, followed by local tuning via \textit{stochastic gradient descent} (SGD), 
%\cite{zhang20}, 
resulting in a high risk of getting stuck in poor-performing objective function valleys. 

An alternative to ``black-box" network architectures is \textit{algorithm unrolling}\,\cite{monga21}, where iterations of an iterative algorithm minimizing a well-defined optimization objective are implemented as a sequence of neural layers to compose a feed-forward network. 
Selected parameters are subsequently optimized end-to-end via back-propagation. 
As one notable example, \cite{yu23nips} unrolled an algorithm minimizing a sparse rate reduction (SRR) objective, resulting in a ``white-box" transformer-like neural net that is ``100\% mathematically interpretable". 
Orthogonally, researchers in \textit{graph signal processing} (GSP)\,\cite{ortega18ieee,cheung18}
recently unrolled graph algorithms minimizing smoothness priors such as \textit{graph Laplacian regularizer} (GLR)\,\cite{pang17} and \textit{graph total variation} (GTV)\,\cite{bai19} into lightweight transformers for image interpolation\,\cite{do2024interpretable}. 
The key insight is that \textit{a graph learning module with normalization is akin to the classical self-attention mechanism} \cite{bahdanau2015neural}, and thus a neural net unrolled from a learnable graph algorithm is technically a transformer. 
However, parameter initialization  for per-pixel feature representation is still done randomly.

\begin{figure}[t]
    \centering
    \includegraphics[width=1\linewidth]{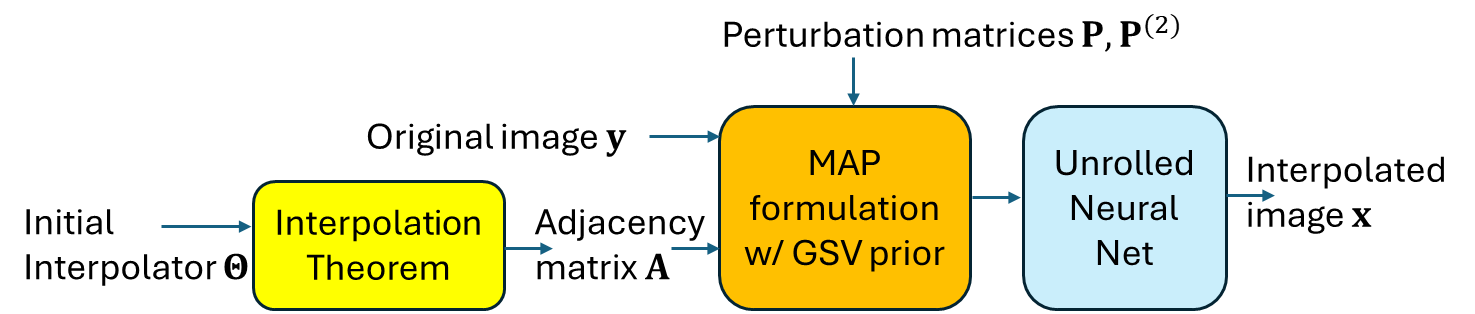}
    \vspace{-0.2in}
    \caption{Overview of Graph Algorithm Unrolling Scheme}
    \label{fig:overview}
    \vspace{-0.1in}
\end{figure}

In this paper, leveraging a recent theorem \cite{Viswarupan2024q} that maps a (pseudo-)linear interpolator $\bTheta$ to a directed graph filter that is a solution to a corresponding  \textit{maximum a posteriori} (MAP) problem  using \textit{graph shift variation} (GSV) \cite{romano17} as signal prior, we first initialize a neural net, unrolled from an iterative graph algorithm solving the MAP problem, to a known interpolator $\bTheta$, so that it has  guaranteed baseline performance.
Then, we augment the initialized adjacency matrix $\A$ in the graph filter in a data-driven manner using two perturbation matrices $\P$ and $\P^{(2)}$ representing two graphs: i) a \textit{directed} graph $\cG^d$ connecting interpolated pixels to observed pixels to improve interpolation, and ii) an \textit{undirected} graph $\cG^u$ inter-connecting interpolated pixels for further denoising.
Finally, we implement the restoration effects of $\P$ and $\P^{(2)}$ progressively via \textit{Douglas-Rachford} (DR) iterations \cite{Giselsson2017}, so that they can be data-tuned individually.
See Fig.\;\ref{fig:overview} for an illustration.
\textit{Compared to \cite{do2024interpretable}, initializing the graph with $\mathbf{A}$ ensures the optimization begins within a high-quality objective function basin; consequently, the learned perturbations $\P$ and $\P^{(2)}$ require lower energy to reach an optimal solution, making the training process significantly less susceptible to gradient instabilities.}

\begin{figure}[t]
    \centering
    \includegraphics[width=1.0\linewidth]{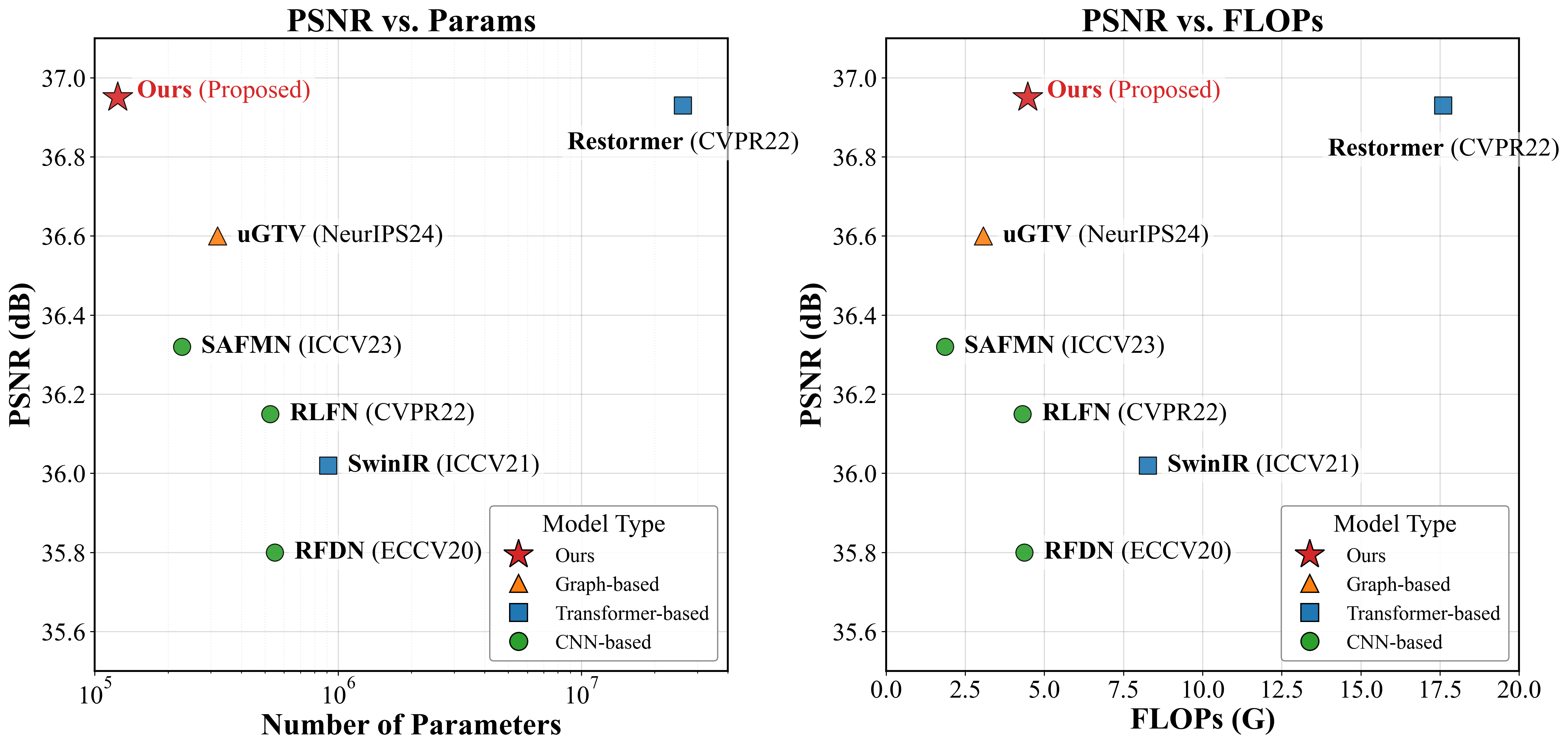}
    \vspace{-0.25in}
    \caption{Tradeoff between PSNR and parameters (left) / FLOPs (right). PSNR denotes the performance on Kodak24~\cite{kodak1993kodak1} under uniform sampling; FLOPs are calculated for a $64 \times 64$ input.}
    \label{fig:psnr_vs_params_vs_plots}
    \vspace{-0.2in}
\end{figure}

Experiments show that our unrolled transformer-like neural net achieves state-of-the-art (SOTA) image interpolation performance compared to SwinIR\,\cite{Liang_2021_ICCV}, Restormer\,\cite{zamir2022restormer} and graph-based uGTV\,\cite{do2024interpretable}, while drastically reducing network parameters (\eg, \textbf{roughly 1\% of parameters in Restormer\,\cite{zamir2022restormer}}) and inference complexity. 
Fig.\;\ref{fig:psnr_vs_params_vs_plots} shows that our model offers the best tradeoffs between image quality and parameter count / FLOPs among SOTA schemes.
Our lightweight model is suitable for real-time applications running on hardware-constrained end-devices, such as unmanned aerial vehicles (UAVs)
%~\cite{bonatti2019towards} 
and autonomous surveillance systems.
%~\cite{zhang2021efficient}.

%\textit{maximum a posteriori} (MAP) problem for image restoration, such as denoising \cite{pang17}, JPEG image dequantization \cite{liu17}, and interpolation \cite{chen24}.

%% file: prelim.tex
\subsection{Graph Shift Variation}

A smooth (consistent) signal $\x \in \mathbb{R}^N$ with respect to (w.r.t.) a graph $\cG$ can be mathematically described in many ways, \eg, \textit{graph Laplacian regularizer} (GLR) \cite{pang17}, \textit{graph total variation} (GTV) \cite{bai19}. 
Here, we focus on \textit{graph shift variation} (GSV) \cite{romano17}, which is well-defined for undirected and directed graphs:
\begin{align}
R(\x) = \| \x - \A \x  \|^2_2 
\label{eq:GSV}
\end{align}
where $\A \in \mathbb{R}^{N \times N}$ is a \textit{graph shift operator} (GSO) \cite{ortega18ieee}.
One example of $\A$ is the row-stochastic adjacency matrix $\D^{-1} \W$, where $\W \in \mathbb{R}^{N \times N}$ is the directed graph adjacency matrix (\ie, $W_{j,i}$ is the weight of an edge connecting nodes $i$ to $j$), and $\D$ is the diagonal \textit{in-degree matrix} (\ie, $D_{i,i} = \sum_j W_{i,j}$).
\eqref{eq:GSV} states that signal $\x$ and its shifted version $\A \x$ should be similar in an $\ell_2$-norm sense. 
%Unlike GLR, GSV can be used for directed graphs also.

\subsection{Interpolation Theorem}

We review the interpolator theorem in \cite{Viswarupan2024q}. 
Consider a linear interpolator $\bTheta \in \mathbb{R}^{N \times M}$ that interpolates $N$ new pixels from $M$ original pixels $\y \in \mathbb{R}^M$.  
Mathematically, we write
\begin{align}
\x = \left[ \begin{array}{c}
\I_{M} \\
\bTheta
\end{array} \right] \y 
\label{eq:interpolator}
\end{align}
where $\x = [\x_M; \x_N] \in \mathbb{R}^{M+N}$ is the length-$(M+N)$ target signal that retains the original $M$ pixels, \ie, $\x_M = \y$.
Note that \eqref{eq:interpolator} includes broadly \textit{pseudo-linear} operator $\bTheta(\y)$ also, where interpolation weights are first computed as (possibly non-linear) functions of input $\y$, then interpolated pixels are computed via matrix-vector multiplication. 
\textit{Bilateral filter} (BF) \cite{tomasi98} is one example of a pseudo-linear operator $\bTheta(\y)$, where the range filter component in BF is computed based on input pixel intensity differences.

Consider next a graph filter that is a solution to a MAP problem with GSV \eqref{eq:GSV} as signal prior. 
%Denote by $\y \in \mathbb{R}^M$ an observation vector containing $M$ observed samples. 
Denote by $\H = [\I_M \; \0_{M,N}] \in \mathbb{R}^{M \times (M+N)}$ a \textit{sampling matrix} that selects the first $M$ original samples from vector $\x$, where $\0_{M,N}$ is a $M \times N$ zero matrix. 
Denote by $\A \in \mathbb{R}^{(M+N) \times (M+N)}$ an \textit{asymmetric} adjacency matrix specifying directional edges in a directed graph $\cG^d$. 
Specifically, $\A$ describes edges only from $N$ new pixels back to the $M$ original pixels, \ie,
\begin{align}
\A= \left[ \begin{array}{cc}
\0_{M,M} & \A_{M,N} \\
\0_{N,M} & \0_{N,N}
\end{array} \right].
\end{align}

We now write an MAP objective using GSV \eqref{eq:GSV} as prior:
\begin{align}
\min_{\x} ||\y-\H \x||_2^2+\mu ||\H(\x-\A\x)||_2^2.
\label{eq:obj}
\end{align}
%In words, GSV states that smooth signal $\x$ and its graph-shifted version $\A \x$ should be similar. 
The two terms in objective \eqref{eq:obj} are both convex for any $\H$ and $\A$. 
To obtain solution $\x^*$ to \eqref{eq:obj}, we take the derivative w.r.t. $\x$ and set it to 0, resulting in 
\begin{align}
\underbrace{\left(\H^{\top}\H + \mu (\I-\A)^{\top}\H^{\top}\H(\I-\A)\right)}_{\C} \x^* = \H^{\top}\y.
\label{eq:linSys}
\end{align}
Given the definitions of $\H$ and $\A$, $\C$ has a unique structure:
\begin{align}
\C = \left[ \begin{array}{cc}
(1+\mu) \I_M & - \mu \A_{M,N} \\
-\mu (\A_{M,N} )^\top & \mu \A_{N,N}^2 
\end{array}
\right] ,
\end{align}
where $\A_{N,N}^2 \triangleq (\A_{M,N})^\top \A_{M,N}$. 
The solution $\x^*$ to \eqref{eq:linSys} is
\begin{align}
\x^* = \C^{-1}\H^{\top}\y
=\left[ \begin{array}{c}
\I_M \\
\A^{-1}_{M,N}
\end{array} \right] \y
\label{eq:interpolated_signal}
\end{align}
where we assume $\A_{M,N}$ is square and invertible, \ie, $M=N$ and $\A_{M,N}$ contains no zero eigenvalues.
%Thus, $[\I_M; \A_{M,N}^{-1}]$ is the solution graph filter to MAP problem \eqref{eq:obj}.

We restate the interpolator theorem in \cite{Viswarupan2024q} to connect a linear interpolator $[\I_M; \bTheta]$ \eqref{eq:interpolator} to a corresponding graph filter that is a solution to the MAP problem \eqref{eq:obj} using GSV as prior.

\noindent
\begin{theorem}
Interpolator $[\I_M;\mathbf{\bTheta}]$ is the solution filter to the
MAP problem \eqref{eq:obj} if $M=N$, $\bTheta$ is invertible, and $\A_{M,N}=\bTheta^{-1}$.
\label{thm:interMapping}
\end{theorem}
%\noindent
%See \cite{Viswarupan2024q} for a proof.

\vspace{0.05in}
\noindent
\textbf{Remark}: One can apply Theorem\;\ref{thm:interMapping} adaptively for a general interpolator that interpolates any $N$ new pixels from $M$ original pixels.
For $N < M$, one can interpolate $M-N$ dummy pixels and then discard them after interpolation.
For $N > M$, one can construct $\lceil \frac{N}{M} \rceil$ interpolators and apply Theorem\;\ref{thm:interMapping} separately.

%% file: formulate.tex
\subsection{Feed-Forward Network Construction}
\label{subsec:ffnet}

Theorem\;\ref{thm:interMapping} states that, under mild conditions, there is a one-to-one mapping from an interpolator $[\I_M; \bTheta]$ to a directed graph filter that is a solution to MAP problem \eqref{eq:obj}.
Leveraging Theorem\;\ref{thm:interMapping}, our goal is to first unroll a graph algorithm minimizing \eqref{eq:obj} to a neural net initialized to $\bTheta$, then further improve performance via data-driven parameter learning.

Specifically, given a linear interpolator $\mathbf{\Theta}$, we leverage Theorem\;\ref{thm:interMapping} and initialize $\A_{M,N}=\bTheta^{-1}$.
We then perturb $\A$ to $\tilde{\A} = \A + \P$ for perturbation matrix $\P$ representing a \textit{directed} graph $\cG^d$ connecting interpolated pixels to original ones:
\begin{align}
\P = \left[ \begin{array}{cc}
\0_{M,M} & \P_{M,N} \\
\0_{N,M} & \0_{N,N}
\end{array} 
\right] .
\label{eq:perturbation_matrix}
\end{align}
From \eqref{eq:interpolated_signal}, it is clear that the new interpolated signal $\x^*$ is

\vspace{-0.1in}
\begin{small}
\begin{align}
\x^* &= \left[ \begin{array}{cc}
\I_M \\
\tilde{\A}_{M,N}^{-1}
\end{array} \right] \y
= \left[ \begin{array}{cc}
\I_M \\
(\A_{M,N} + \P_{M,N})^{-1}
\end{array} \right] \y .
\label{eq:interpolate2}
\end{align}
\end{small}\noindent
However, \eqref{eq:interpolate2} involves a computation-intensive matrix inversion, with worst-case complexity $\cO(N^3)$.

\vspace{0.05in}
\subsubsection{Computing Interpolated Pixels}

Given interpolator $\bTheta$, to compute interpolated pixels $\x_N^* \in \mathbb{R}^N$ \textit{without} matrix inversion, we rewrite \eqref{eq:interpolate2} as
\begin{align}
(\A_{M,N} + \P_{M,N}) \x_N^* &= \y
\label{eq:linSys1} \\
\x_N^* + \underbrace{\A_{M,N}^{-1}}_{\bTheta} \P_{M,N} \x_N^* &= \underbrace{\A_{M,N}^{-1}}_{\bTheta} \y
\\
(\I_N + \bTheta \P_{M,N}) \x_N^* &= \bTheta \, \y .
\label{eq:interpolate3}
\end{align}
\eqref{eq:interpolate3} is a linear system, and though the coefficient matrix $\I_M + \bTheta \P_{M,N}$ is not symmetric in general, $\x_N^*$ can nonetheless be computed efficiently via \textit{biconjugate gradient} (BiCG) \cite{1992Bi}---an iterative algorithm similar to \textit{conjugate gradient} (CG) \cite{1952Methods}---extended to asymmetric coefficient matrices.

\vspace{0.05in}
\noindent
\textbf{Complexity}: 
The complexity of BiCG to solve linear system $\Q \x = \y$ is $\cO(\text{nnz}(\Q) \sqrt{\kappa(\Q)} / \log(\epsilon) )$, where $\text{nnz}(\Q)$ is the number of non-zero entries in coefficient matrix $\Q$, $\kappa(\Q) = \frac{\lambda_{\max}(\Q)}{\lambda_{\min}(\Q)}$ is the condition number of $\Q$, and $\epsilon$ is the chosen convergence threshold to terminate the iterative algorithm.
Assuming $\kappa(\I_N + \bTheta \P_{M,N})$ and $\epsilon$ have reasonable values, the complexity of BiCG is $\cO(N^2)$ for dense $\bTheta \P_{M,N}$.  

%Thus, given $\bTheta$, $\x_N^*$ in \eqref{eq:interpolate3} can be computed without any matrix inversion operations. 

\vspace{0.05in}
We unroll BiCG iterations into neural layers to compose a feed-forward network as done in \cite{do2024interpretable}, so that parameters for BiCG and $\P_{M,N}$ can be optimized in a data-driven manner. 
Specifically, each BiCG iteration has two parameters, $\alpha_{\text{BiCG}}$ and $\beta_{\text{BiCG}}$, that correspond to step size and momentum term respectively.
$\P_{M,N}$ specifies an $N$-node directed graph $\cG^d$, where \textit{signed} edge weights $w_{i,j}\in [-1,1]$ can be computed as a function of \textit{feature distance} $d_{i,j}$:

\vspace{-0.1in}
\begin{small}
\begin{align}
w_{i,j} = \frac{-2}{1+e^{ -\left( d_{i,j}-d^* \right)}}+1, 
~~
d_{i,j} = (\f_i - \f_j)^\top \M (\f_i - \f_j) .
\label{eq:signed_edge}
\end{align}
\end{small}\noindent
$\f_i \in \mathbb{R}^K$ is a $K$-dimensional feature vector for pixel $i$, computed from data via a shallow CNN \cite{do2024interpretable}. 
$\M \succeq 0 \in \mathbb{R}^{K \times K}$ is a PSD \textit{metric matrix}, also learned from data, and $d^* \gg 0$ is a parameter.
By \eqref{eq:signed_edge},  larger feature distance means smaller edge weights, and thus $\cG^d$ is a \textit{similarity} graph.
Signed edges are important, so that directed edge weights in $\A_{M,N}$ can be augmented in both directions into $\A_{M,N} + \P_{M,N}$.

\vspace{0.05in}
\subsubsection{Cascading Filter Interpretation}

Towards an intuitive interpretation, we rewrite \eqref{eq:interpolate3} as 
\begin{align}
\x_N^* &= \underbrace{(\I_M + \bTheta \P_{M,N})^{-1}}_{\bTheta_{P}} \bTheta \, \y = \bTheta_{P} \, \bTheta \, \y. 
\label{eq:interpolate3b}
\end{align}
We see that using perturbation matrix $\P_{M,N}$ to augment adjacency matrix $\A_{M,N}$ is equivalent to filtering the original interpolated output $\bTheta \y$ \textit{again} using a new filter $\bTheta_{P}$. 
In one special case when $\P_{M,N} = \0_{M,N}$, $\bTheta_{P} = \I_M$, as expected.
In another special case when $\P_{M,N} = \A_{M,N}$, $\bTheta_{P} = (\I_M + \bTheta \A_{M,N})^{-1} = (\I_M + \I_M)^{-1} = \frac{1}{2} \I_M$. This also makes sense; in this case the perturbed adjacency matrix is $\tilde{\A}_{M,N} = 2 \A_{M,N}$, and because of the matrix inverse relationship between $\bTheta$ and $\A_{M,N}$, the equivalent filter in the pixel domain is $\bTheta_{P} \bTheta = \frac{1}{2}\bTheta$.

\vspace{0.05in}
\subsubsection{Multiple Perturbations via Cascades}

Can we leverage the cascading filter interpretation in \eqref{eq:interpolate3b} to implement multiple perturbations? 
Considering $\tilde{\A} = \A + \P$ as the original adjacency matrix with corresponding filter $\tilde{\bTheta} = \bTheta_{P} \bTheta$, we can perturb it again with $\P^{(2)}$, resulting in $\tilde{\A}^{(2)} = \tilde{\A} + \P^{(2)} = \A + \P + \P^{(2)}$. 
The interpolated signal $\x^{(2)}_N$ is then
\begin{align}
\x^{(2)}_N &= (\I_M + \tilde{\bTheta} \P^{(2)}_{M,N})^{-1} \tilde{\bTheta} \, \y
\nonumber \\
(\I_M + \bTheta_{P} \bTheta \P^{(2)}_{M,N}) \x^{(2)}_N &= \x^*_N = \bTheta_{P} \, \bTheta \, \y .
\label{eq:interpolate_again}
\end{align}
%Thus, \eqref{eq:interpolate3b} provides one mechanism to cascade filters in sequence. 
However, to compute linear system \eqref{eq:interpolate_again}, one must first compute matrix inverse $\bTheta_{P} \triangleq (\I_M + \bTheta \P_{M,N})^{-1}$, which has $\cO(N^3)$ complexity.
We seek a more computation-efficient alternative in the sequel.

\subsection{Second Perturbation Matrix via Denoising}

Revisiting the original MAP optimization \eqref{eq:obj}, it is apparent that the objective is a combination of two $\ell_2$-norm terms: 
\begin{align}
\min_{\x} & \|\y - \H \x\|^2_2 + \mu \|\H (\I - \A) \x\|^2_2 
\nonumber \\
=& \|\y - \H \x\|^2_2 + \mu  \x^\top \underbrace{(\I - \A)^\top \H^\top \H (\I - \A)}_{\B} \x .
\end{align}
We can perturb the symmetric and PSD matrix $\B$ with another symmetric and PSD Laplacian matrix $\P^{(2)} \in \mathbb{R}^{N \times N}$, corresponding to an \textit{undirected} positive graph\footnote{A positive graph $\cG$ is a graph with only non-negative edge weights.} $\cG^u$ inter-connecting \textit{only} the $N$ interpolated pixels for denoising, results in
\begin{align}
=& \underbrace{\|\y - \H \x\|^2_2 + \mu \,  \x^\top \B \x}_{h(\x)} + \underbrace{\mu \, \x^\top \G^\top \P^{(2)} \G \x}_{g(\x)} ,
\label{eq:obj2}
\end{align}
where $\G = [\0_{N,M} \; \I_N] \in \mathbb{R}^{N \times (M+N)}$, a complementary sampling matrix to $\H$, selects only the $N$ interpolated pixels from $\x$.
The objective \eqref{eq:obj2} is a sum of two functions: i) $h(\x)$ composed of the fidelity term and a GSV prior involving $\tilde{\A}$, and ii) $g(\x)$ that is a GLR prior involving $\P^{(2)}$. 

To minimize \eqref{eq:obj2} while reusing the earlier mathematical development, we first recall that one expression of the \textit{Douglas-Rachford} (DR) splitting method (eq. (10) to (12) in \cite{Giselsson2017}) is 
\begin{align}
\z(k) &= \text{prox}_{\gamma h}(\x(k)) 
\label{eq:DR1} \\
\v(k) &= \text{prox}_{\gamma g}(2 \z(k) - \x(k))
\label{eq:DR2} \\
\x(k+1) &= \x(k) + 2\alpha(\v(k) - \z(k)) ,
\label{eq:DR3}
\end{align}
where $\alpha \in (0,1)$, and $\text{prox}_{\gamma h}(\x(k))$ is the proximal mapping of $h(\cdot)$ for argument $\x(k)$ with $\gamma \in (0,1)$.
We design an iterative optimization method based on DR to compute proximal mappings of $h(\x)$ and $g(\x)$ alternately until convergence:

\vspace{0.05in}
\noindent
\textbf{$h$-step}: 
We compute the proximal mapping of $h(\cdot)$ with argument $\x_N(k)$ for $\z_N(k)$ at iteration $k$:

\vspace{-0.1in}
\begin{small}
\begin{align}
(\I_N + \bTheta \P_{M,N}) \z_N(k) &= \bTheta (\y + \frac{1}{2\gamma} (\x_N(k-1) - \x_N(k)) .
\label{eq:prox1}
\end{align}
\end{small}\noindent
Note that we compute $\z_N(k)$ in \eqref{eq:prox1} via BiCG without matrix inversion, similar to \eqref{eq:interpolate3}.
See Appendix\;\ref{sec:append1} for a derivation. 

\vspace{0.05in}
\noindent
\textbf{$g$-step}: 
Given $\z_N(k)$, we compute the proximal mapping of $g(\cdot)$ with argument $2 \z_N(k) - \x_N(k)$ for $\v_N(k)$.
Solution is
\begin{align}
(2\mu \P^{(2)} + \gamma^{-1} \I_N) \v_N(k) = \gamma^{-1} \left( 2 \z_N(k) - \x_N(k) \right) . 
\label{eq:prox2}
\end{align}
See Appendix\;\ref{sec:append2} for a derivation.

Given $\x_N(k)$, computed $\z_N(k)$ and $\v_N(k)$, $\x_N(k+1)$ for iteration $k+1$ is updated using \eqref{eq:DR3}. 
$h$-step, $g$-step, and $\x$-update are repeated until convergence.

%% file: unrolling.tex
\begin{figure}[t]
    \centering
    \includegraphics[width=1\linewidth]{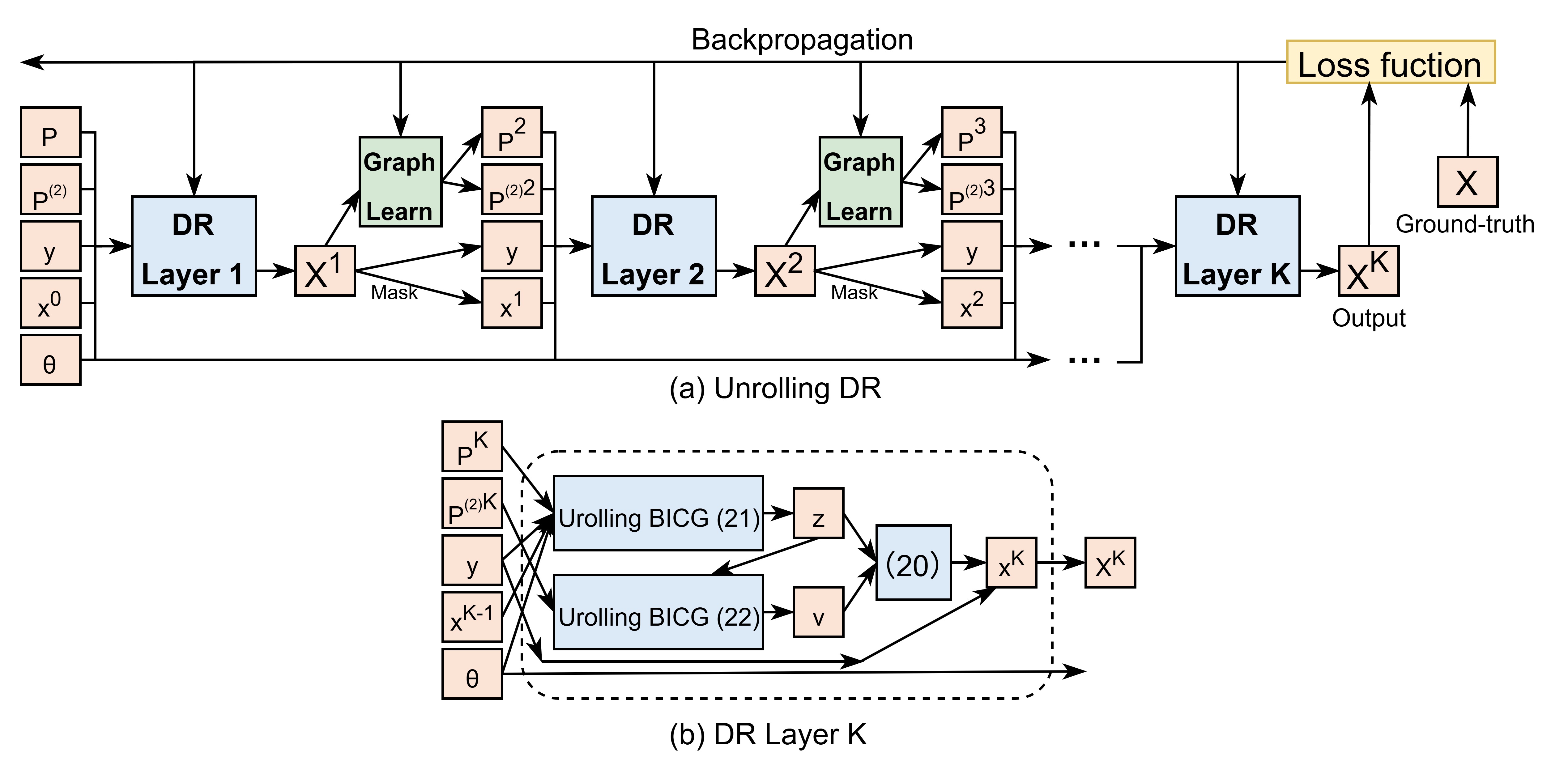}
    \vspace{-0.25in}
    \caption{Network from Unrolling of DR iterations into neural layers. Note that a graph learning module (self-attention mechanism) is periodically inserted, as done in \cite{do2024interpretable}.}
    \label{fig:architecture}
    \vspace{-0.1in}
\end{figure}

We unroll 15 DR iterations into corresponding neural layers to compose a feed-forward network, so that chosen parameters can be tuned using data.
As done in \cite{do2024interpretable}, a graph learning module is periodically inserted; see Fig.\,\ref{fig:architecture} for an illustration. 
For each DR layer, feature vectors are learned to compute feature distances, edge weights, and thus perturbation matrices $\P$ and $\P^{(2)}$. 
Unlike \eqref{eq:signed_edge} used for $\P$,  $\P^{(2)}$ specifies an undirected positive graph $\cG^u$ with positive edge weight $w'_{i,j}$ and feature distance $d'_{i,j}$ defined as
\begin{align}
w'_{i,j} = \exp \left( - d'_{i,j} \right), 
~~~
d'_{i,j} = (\f'_i - \f'_j)^\top \R (\f'_i-\f'_j).
\label{eq:unsigned_edge}
\end{align}
$\f_i' \in \mathbb{R}^K$ is a feature vector for pixel $i$, computed from data via a shallow CNN. 
$\R \succeq 0 \in \mathbb{R}^{K \times K}$ is a PSD metric matrix learned from data.
This ensures $w'_{i,j} \in (0,1]$ and $\P^{(2)}$ is PSD.

%\red{need more text to describe what parameters are learned in what layers.}

Inside each DR layer, the two linear systems for the $h$-step \eqref{eq:prox1} and $g$-step \eqref{eq:prox2} are solved by two separate unrollings of BiCG, each with 5 layers. 
Step size $\alpha_{\text{BiCG}}$ and momentum $\beta_{\text{BiCG}}$ in each BiCG layer are tuned. 
In summary, the learned network parameters are i) scalar parameters $\alpha$, $\gamma$ and $\mu$ governing the DR iterations; ii) weights of the CNNs that compute feature vectors $\f_i$'s and $\f_i'$'s, along with the metric matrices $\M$ and $\R$ to construct $\P$ and $\P^{(2)}$; and iii) the aforementioned BiCG parameters  $\alpha_{\text{BiCG}}$ and $\beta_{\text{BiCG}}$.

%% file: results.tex
\subsection{Experimental Setup}
\label{subsec:setup}

All models were implemented in PyTorch (Python 3.12) and trained on an NVIDIA GeForce RTX 3090 GPU.
To train each learning model, we used the DIV2K\,\cite{agustsson2017ntire} dataset, containing $800$ high-res training images and $100$ validation images. 
Each training image was segmented into 10 non-overlapping $64 \times 64$ patches.
For evaluation, we used the Kodak\,\cite{kodak1993kodak1}, Urban100\,\cite{huang2015single}, and McM\,\cite{zhang2011color} datasets. 
%Owing to the model's architectural design, 
Note that our method operated on $64 \times 64$ patches with a 16-pixel overlap for interpolation, and overlapped regions were averaged to achieve smooth image reconstruction. 
In contrast, competing methods performed interpolation directly on the full image.

We focused on image interpolation in the single-channel Y (luminance). 
To create input, we considered two sampling scenarios to simulate pixel loss: i) \textit{Uniform Sampling}, where we removed 50\% pixels from a full Y-channel image in a regular checkerboard pattern; and ii) \textit{Random Sampling}, where 50\% pixels were removed randomly from the image grid.
We employed a learnable version of the traditional Bicubic interpolator, named \texttt{Bicubic+}, as the initial pseudo-linear interpolator $\mathbf{\Theta}$ for our model. 
Unlike conventional Bicubic which employs fixed analytic kernels, \texttt{Bicubic+} utilizes a lightweight MLP to predict the $4\times4$ weights based on coordinate differences.
This enables adaptation to local image content.
This initial interpolator $\bTheta$ ensured the subsequent unrolled network parameter optimization started within a high-quality objective function basin. 

As illustrated in Fig.~\ref{fig:overview}, we first obtain the initial interpolated image via the \texttt{Bicubic+} interpolator and the adjacency matrix $\mathbf{A}$ using the interpolation theorem. Subsequently, our model employs a two-stage cascaded perturbation network ($\mathbf{P} + \mathbf{P}^{(2)}$), where the perturbation matrices are generated by a graph learning module featuring a shallow CNN feature extractor with 2 convolutional layers, each producing 48 feature maps. 
The architecture consists of 15 unfolded DR layers (see Fig.~\ref{fig:architecture}), each containing 5 BiCG layers. Training utilized the Adam optimizer with a batch size of 8, a learning rate of 1e-3, and an early stopping patience of 5.

%To comprehensively evaluate the performance of the proposed two-stage cascaded restoration network, 
We considered seven representative competing methods: model-based method (Bicubic interpolation), CNN-based methods (RFDN \cite{liu2020residual}, RLFN \cite{kong2022residual}, SAFMN \cite{sun2023spatially}), transformer-based methods (SwinIR \cite{Liang_2021_ICCV}, Restormer \cite{zamir2022restormer}), and graph-based method (uGTV \cite{do2024interpretable}). 
%These baselines encompass the mainstream technical approaches in the current image restoration field. 
We evaluated performance on test images using two common image metrics: Peak Signal-to-Noise Ratio (PSNR) and Structural Similarity Index (SSIM) \cite{wang04}.

\subsection{Experimental Results}
\label{subsec:results}

% 表1 单栏
\begin{table}[t]
\centering
%\captionsetup{font={small,bf}, justification=centering}
\caption{Quantitative comparison under uniform and random sampling scenarios. The best and second best results are indicated in \textbf{bold} and \underline{underline}.}
\vspace{-0.1in}
\resizebox{\columnwidth}{!}{%
\begin{tabular}{@{}ccccc@{}}
\toprule
\multirow{2}{*}{\textbf{Method}} &
\multirow{2}{*}{\textbf{Params}} &
\textbf{Kodak\cite{kodak1993kodak1}} &
\textbf{Urban100\cite{huang2015single}} &
\textbf{McM\cite{zhang2011color}} \\
& & \textbf{PSNR $\vert$ SSIM} & \textbf{PSNR $\vert$ SSIM} & \textbf{PSNR $\vert$ SSIM} \\
\midrule
% --- 第一部分：均匀采样 (Uniform/Grid) ---
\multicolumn{5}{c}{\textbf{\textit{Setting 1: Uniform Sampling (Grid)}}} \\
\midrule
Bicubic
& -- & 32.59 $\vert$ 0.9332 & 28.38 $\vert$ 0.9176 & 34.88 $\vert$ 0.9578 \\
%SRCNN\,\cite{7115171} & 57,281 & 35.25 $\vert$ 0.9599 & 31.00 $\vert$ 0.9478 & 39.02 $\vert$ 0.9786 \\
RFDN\,\cite{liu2020residual}
& 533,568 & 35.80 $\vert$ 0.9605 & 31.42 $\vert$ 0.9501 & 39.45 $\vert$ 0.9792 \\
RLFN\,\cite{kong2022residual}
& 526,465 & 36.15 $\vert$ 0.9625 & 31.75 $\vert$ 0.9535 & 40.05 $\vert$ 0.9810 \\
SAFMN\,\cite{sun2023spatially}
& 227,624 & 36.32 $\vert$ 0.9630 & 32.05 $\vert$ 0.9560 & 40.20 $\vert$ 0.9813 \\
SwinIR-Light\,\cite{Liang_2021_ICCV}
& 910,152 & 36.06 $\vert$ 0.9619 & 31.69 $\vert$ 0.9525 & 39.93 $\vert$ 0.9809 \\
Restormer\,\cite{zamir2022restormer}
& 26,124,052 & \underline{36.93 $\vert$ 0.9656} & \textbf{33.97 $\vert$ 0.9657} & \textbf{40.92 $\vert$ 0.9825} \\
uGTV\,\cite{do2024interpretable}
& 319,115 & 36.73 $\vert$ 0.9653 & 32.92 $\vert$ 0.9610 & 40.43 $\vert$ 0.9818 \\
Ours (Bicubic) & \textbf{159,130} & 36.61 $\vert$ 0.9658 & 32.86 $\vert$ 0.9610 & 40.35 $\vert$ 0.9819 \\
Ours (Bicubic+) & 164,092 & \textbf{36.98 $\vert$ 0.9668} & \underline{33.55 $\vert$ 0.9642} & \underline{40.89 $\vert$ 0.9833} \\

\midrule
% --- 第二部分：随机采样 (Random Sampling) ---
\multicolumn{5}{c}{\textbf{\textit{Setting 2: Random Sampling (50\%)}}} \\
\midrule
Bicubic
& -- & 30.76 $\vert$ 0.9078 & 26.67 $\vert$ 0.8859 & 32.91 $\vert$ 0.9394 \\
%SRCNN\,\cite{7115171} & 57,281   & 30.89 $\vert$ 0.9197 & 27.23 $\vert$ 0.9008 & 33.91 $\vert$ 0.9464 \\
RFDN\,\cite{liu2020residual}
& 533,568 & 32.12 $\vert$ 0.9345 & 28.85 $\vert$ 0.9250 & 35.68 $\vert$ 0.9610 \\
RLFN\,\cite{kong2022residual}
& 526,465 & 33.40 $\vert$ 0.9390 & 29.65 $\vert$ 0.9310 & 36.50 $\vert$ 0.9635 \\
SAFMN\,\cite{sun2023spatially}
& 227,624 & 33.90 $\vert$ 0.9420 & 30.15 $\vert$ 0.9380 & 36.85 $\vert$ 0.9650 \\
SwinIR-Light\,\cite{Liang_2021_ICCV}
& 910,152  & 32.26 $\vert$ 0.9309 & 28.41 $\vert$ 0.9144 & 35.45 $\vert$ 0.9582 \\
Restormer\,\cite{zamir2022restormer}
& 26,124,052 & \underline{34.25 $\vert$ 0.9443} & \textbf{31.40 $\vert$ 0.9470} & \underline{37.61 $\vert$ 0.9679} \\
uGTV\,\cite{do2024interpretable}
& 319,115  & 34.11 $\vert$ 0.9440 & 30.35 $\vert$ 0.9423 & 37.12 $\vert$ 0.9672 \\
Ours (Bicubic) & \textbf{159,130} & 34.23 $\vert$ 0.9442 & 30.92 $\vert$ 0.9451 & 37.41 $\vert$ 0.9672 \\
Ours (Bicubic+) & 164,092 & \textbf{34.38 $\vert$ 0.9461} & \underline{31.39 $\vert$ 0.9470} & \textbf{37.75 $\vert$ 0.9681} \\

\bottomrule
\end{tabular}%
}
\label{tab:main_results}
\end{table}

%表2 单栏
\begin{table}[t]
\centering
%\captionsetup{font={small,bf}, justification=centering} 
\caption{Quantitative results of different configurations evaluated on 100 validation images from the DIV2K dataset.}
\vspace{-0.1in}
\resizebox{\columnwidth}{!}{%
\begin{tabular}{@{}ccccccc@{}}
\toprule
\multirow{2}{*}{\textbf{Configuration}} &
\multirow{2}{*}{\textbf{$\bTheta$ Type}} & 
\multirow{2}{*}{\textbf{Matrix $\P$}} & 
\multirow{2}{*}{\textbf{$\P$ Weights}} & 
\multirow{2}{*}{\textbf{Matrix $\P^{(2)}$ }} & 
\multirow{2}{*}{\textbf{PSNR (dB)}} & 
\multirow{2}{*}{\textbf{$\Delta$PSNR (dB)}} \\ & & & & & \\
\midrule
Baseline & Bicubic & $\times$ & -- & $\times$ & 29.02 & -- \\
Variant 1 & Bicubic+ & $\times$ & -- & $\times$ & 32.48 & +3.46 \\
Variant 2 & Bicubic+ & $\checkmark$ & Positive & $\times$ & 33.53 & +4.51 \\
Variant 3 & Bicubic+ & $\checkmark$ & Signed & $\times$ & 34.76 & +5.74 \\
\textbf{Proposed}  & Bicubic+ & $\checkmark$ & Signed & $\checkmark$ & 35.39 & +6.37 \\
\bottomrule
\end{tabular}%
}
\label{tab:ablation}
\end{table}

\label{sec:results}
\begin{figure*}[t]
    \centering
    \includegraphics[width=1.0\textwidth]{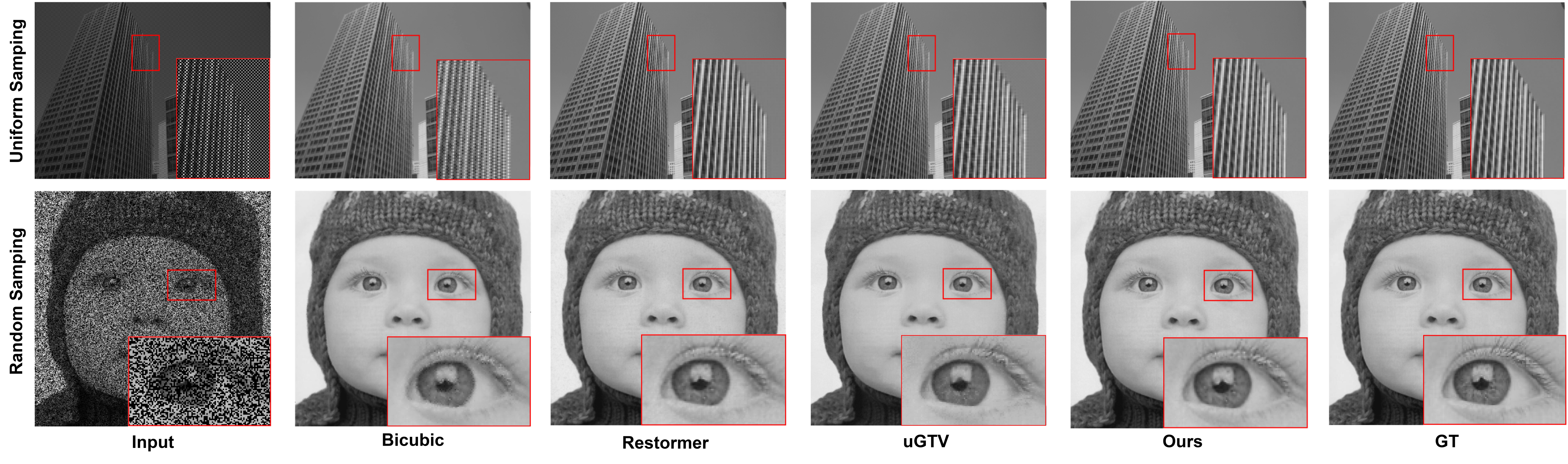}
    \vspace{-0.3in}
   \caption{\small Visual comparison on ``Urban96'' from Urban100\cite{huang2015single} (top, Uniform Sampling) and ``Baby'' from Set5\cite{bevilacqua2012low} (bottom, Random Sampling). Our proposed method achieves visual quality comparable to state-of-the-art methods with \text{significantly reduced parameters}.} 
   %\textbf{Please zoom in for better viewing.}}
    \label{fig:visual}
    \vspace{-0.2in}
\end{figure*}

As shown in Table\,\ref{tab:main_results}, our proposed model achieves near-identical performance to state-of-the-art (SOTA) Restormer, while using significantly fewer parameters. 
Specifically, our model employs only about 160k parameters---\textbf{approximately 0.6\% of Restormer}---while maintaining competitive PSNR/SSIM scores across all three datasets. 
Compared to other lightweight methods (e.g., uGTV, SAFMN), our method demonstrates noticeably better performance under both uniform and random sampling scenarios.

The tradeoffs between reconstructed image quality (in PSNR) and parameter count / FLOPs are shown in Fig.\;\ref{fig:psnr_vs_params_vs_plots}. 
Specifically, we see that in Fig.\;\ref{fig:psnr_vs_params_vs_plots}(left) that our model offers the lowest parameter count while achieving the higher image quality; in Fig.\;\ref{fig:psnr_vs_params_vs_plots}(right), we see that our model has $4.47G$ FLOPs to Restormer's $17.6G$ FLOPs---\textbf{roughly a $75\%$ reduction in inference complexity}. 

A key advantage of our approach is the guaranteed initialization derived from Theorem 1. As shown in Table\,\ref{tab:main_results}, we outperform uGTV (which relies on random initialization) despite using fewer parameters. This confirms that initializing the unrolled network with a known interpolator $\bTheta$ ensures optimization begins within a high-quality basin, avoiding poor local minima. 
Further, the performance gain of ``Ours (Bicubic+)'' over ``Ours (Bicubic)'' confirms the critical benefit of employing an adaptive initializer.

To evaluate the contributions of each component in the proposed framework, we conducted an ablation study on 100 validation images from the DIV2K dataset, as detailed in Table\,\ref{tab:ablation}. Replacing the conventional Bicubic with the adaptive \texttt{Bicubic+} (Variant 1) yields a substantial +3.46 dB gain, validating the importance of a data-driven content-adaptive initializer. The subsequent inclusion of graph perturbations $\P$ (Variant 3) and $\P^{(2)}$ (Proposed) further boosts performance by +2.91 dB. 
These results confirm the importance of our novel dual-perturbation design that improves interpolation from initialized $\A$ and denoises subsequently.

% From Fig.\,\ref{fig:visual}, we observe that \texttt{Bicubic}  exhibits noticeable blurring and detail loss. While Restormer and uGTV improve structure, they still struggle with fine texture restoration. In contrast, our method, initialized by \texttt{Bicubic+} and refined by graph perturbations, achieves the best balance between structural preservation and high-frequency detail recovery.
From Fig.\,\ref{fig:visual}, Bicubic exhibits noticeable blurring. While uGTV improves structure, it misses fine textures. Notably, our method achieves visual quality comparable to the large-size Restormer and superior to the lightweight uGTV, effectively recovering high-frequency details with significantly fewer parameters.

%% file: conclude.tex
\vspace{-0.05in}
Leveraging a recent theorem \cite{Viswarupan2024q} that maps a pseudo-linear interpolator $\bTheta$ to a directed graph solution filter corresponding to a MAP problem using the graph shift variation (GSV) \cite{romano17} as prior, we build a lightweight neural net by unrolling an iterative Douglas-Rachford algorithm \cite{Giselsson2017} solving the MAP problem.
The key novelty is that we can initialize the network as $\bTheta$, thus guaranteeing a baseline performance level, before data-tuning two perturbation matrices corresponding to directed and undirected graphs for further gain. 
Experiments show SOTA interpolation results while reducing parameters drastically.

%% file: append.tex
\subsection{Proof of \eqref{eq:prox1}}
\label{sec:append1}

\vspace{-0.05in}
The proximal mapping for $h(\x(k))$ is
\begin{align}
\min_\x \|\y - \H \x\|^2_2 + \mu \, \x^\top \B \x + \frac{1}{2\gamma} \|\x - \x(k)\|^2_2
\end{align}
Taking the derivative w.r.t. $\x$ and set it to zero, we get
\begin{align}
%- 2 \H^\top \y + 2 \H^\top \H \x^* + 2 \mu \B \x^* - \gamma^{-1} \x_k + \gamma^{-1} \x^* = 0
%\nonumber \\
(2 \H^\top \H + 2 \mu \B + \gamma^{-1} \I) \x^* &= 2 \H^\top \y + \gamma^{-1} \x(k) 
\nonumber \\
\underbrace{(\H^\top \H + \mu \B)}_{\C} \x^* &= \H^\top \y + \frac{1}{2 \gamma} (\x(k) - \x^*)
\end{align}
Assuming $\x(k) \rightarrow \x^*$, we approximate $\x(k) - \x^*$ as $\x(k-1) - \x(k)$:

\vspace{-0.1in}
\begin{small}
\begin{align}
\C \x^* &= \H^\top \y + \frac{1}{2\gamma} (\x(k-1) - \x(k))
\nonumber \\
(\A_{M,N} + \P_{M,N}) \x^*_N &= \y + \frac{1}{2\gamma} (\x_N(k-1) - \x_N(k)) 
\nonumber \\
(\I_N + \bTheta \P_{M,N}) \x_N^* &= \bTheta (\y + \frac{1}{2\gamma} (\x_N(k-1) - \x_N(k)) 
\end{align}
\end{small}\noindent
where the last two lines follow the same derivation as \eqref{eq:interpolate3}.

\subsection{Proof of \eqref{eq:prox2}}
\label{sec:append2}

\vspace{-0.05in}
Denote by $\u(k) \triangleq 2 \z(k) - \x(k)$. 
By definition of $g(\cdot)$, $\text{prox}_{\gamma g}(\u(k))$ is the argument that minimizes
\begin{align}
\min_{\x} ~ \mu \x^\top \G^\top \P^{(2)} \G \x + \frac{1}{2\gamma} \|\x - \u(k) \|^2_2 .
\end{align}
We take the derivative w.r.t. $\x$ and set it to zero:
\begin{align}
2 \mu \G^\top \P^{(2)} \G \x^* - \gamma^{-1} \u(k) + \gamma^{-1} \x^* = 0
\nonumber \\
(2\mu \P^{(2)} + \gamma^{-1} \I_N) \x_N^* = \gamma^{-1} \u_N(k) .
\end{align}

%% file: ref2.bib
@inproceedings{huang2015single,
  title={Single image super-resolution from transformed self-exemplars},
  author={Huang, Jia-Bin and Singh, Abhishek and Ahuja, Narendra},
  booktitle={Proceedings of the IEEE conference on computer vision and pattern recognition},
  pages={5197--5206},
  year={2015}
}

@article{kodak1993kodak1,
  title={Kodak lossless true color image suite (photocd pcd0992)},
  author={{Eastman Kodak Company}},
  journal={URL http://r0k. us/graphics/kodak},
  volume={6},
  pages={2},
  year={1993}
}

@article{zhang2011color,
  title={Color demosaicking by local directional interpolation and nonlocal adaptive thresholding},
  author={Zhang, Lei and Wu, Xiaolin and Buades, Antoni and Li, Xin},
  journal={Journal of Electronic imaging},
  volume={20},
  number={2},
  pages={023016--023016},
  year={2011},
  publisher={Society of Photo-Optical Instrumentation Engineers}
}

@techreport{shewchuk94q,
  author      = {Shewchuk, Jonathan Richard},
  title       = {An Introduction to the Conjugate Gradient Method Without the Agonizing Pain},
  institution = {Carnegie Mellon University},
  year        = {1994},
  address     = {Pittsburgh, PA, USA},
  month       = {Aug}
}

@INPROCEEDINGS(tomasi98,
  TITLE = "Bilateral Filtering for Gray and Color Images",
  AUTHOR = "C. Tomasi and R. Manduchi",
  BOOKTITLE = "Proceedings of the IEEE International Conference on Computer Vision",
  ADDRESS = "Bombay, India",
  YEAR = "1998")

@ARTICLE{wang04,
  author={Zhou Wang and Bovik, A.C. and Sheikh, H.R. and Simoncelli, E.P.},
  journal={IEEE Transactions on Image Processing}, 
  title={Image quality assessment: from error visibility to structural similarity}, 
  year={2004},
  volume={13},
  number={4},
  pages={600-612},
  doi={10.1109/TIP.2003.819861}}

@inproceedings{bahdanau2015neural,
  title={Neural Machine Translation by Jointly Learning to Align and Translate},
  author={Bahdanau, Dzmitry and Cho, Kyunghyun and Bengio, Yoshua},
  booktitle={3rd International Conference on Learning Representations (ICLR)},
  year={2015},
  address={San Diego, CA, USA}
}

@ARTICLE{Giselsson2017,
  author={Giselsson, Pontus and Boyd, Stephen},
  journal={IEEE Transactions on Automatic Control}, 
  title={Linear Convergence and Metric Selection for {Douglas-Rachford} Splitting and {ADMM}}, 
  year={2017},
  volume={62},
  number={2},
  pages={532-544},
  doi={10.1109/TAC.2016.2564160}}

@ARTICLE{pang17,
  author={Pang, Jiahao and Cheung, Gene},
  journal={IEEE Transactions on Image Processing}, 
  title={Graph {Laplacian} Regularization for Image Denoising: Analysis in the Continuous Domain}, 
  year={2017},
  volume={26},
  number={4},
  pages={1770-1785},
  doi={10.1109/TIP.2017.2651400}}

@INPROCEEDINGS(romano17,
  TITLE = "The Little Engine That Could: Regularization by Denoising ({RED})",
  AUTHOR = "Yaniv Romano and Michael Elad and Peyman Milanfar",
  BOOKTITLE = "SIAM Journal on Imaging Sciences",
  VOLUME = "10, no.4",
  PAGES = "1804-1844",
  YEAR = "2017")

@article{vaswani17attention,
  title={Attention is all you need},
  author={Vaswani, Ashish and Shazeer, Noam and Parmar, Niki and Uszkoreit, Jakob and Jones, Llion and Gomez, Aidan N and Kaiser, Lukasz and Polosukhin, Illia},
  journal={Advances in neural information processing systems},
  volume={30},
  year={2017}
}

@INPROCEEDINGS(cheung18,
  TITLE = "Graph Spectral Image Processing",
  AUTHOR = "G. Cheung and E. Magli and Y. Tanaka and M. Ng",
  BOOKTITLE = "Proceedings of the {IEEE}",
  VOLUME = "106, no.5",
  PAGES = "907-930", 
  MONTH = "May",
  YEAR = "2018")

@INPROCEEDINGS(ortega18ieee,
  TITLE = "Graph Signal Processing: Overview, Challenges, and Applications",
  AUTHOR = "A. Ortega and P. Frossard and J. Kovacevic and J. M. F. Moura and P. Vandergheynst",
  BOOKTITLE = "Proceedings of the {IEEE}",
  VOLUME = "106, no.5",
  PAGES = "808-828", 
  MONTH = "May",
  YEAR = "2018")

@article{bai19,
  TITLE = "Graph-Based Blind Image Deblurring from a Single Photograph",
  AUTHOR = "Y. Bai and G. Cheung and X. Liu and W. Gao",
   journal = "IEEE Transactions on Image Processing",
  VOLUME = "28, no.3",
  PAGES = "1404-1418", 
  YEAR = "2019"}

@ARTICLE{monga21,
  author={Monga, Vishal and Li, Yuelong and Eldar, Yonina C.},
  journal={IEEE Signal Processing Magazine}, 
  title={Algorithm Unrolling: Interpretable, Efficient Deep Learning for Signal and Image Processing}, 
  year={2021},
  volume={38},
  number={2},
  pages={18-44},
  doi={10.1109/MSP.2020.3016905}}

@inproceedings{yu23nips,
 author = {Yu, Yaodong and Buchanan, Sam and Pai, Druv and Chu, Tianzhe and Wu, Ziyang and Tong, Shengbang and Haeffele, Benjamin and Ma, Yi},
 booktitle = {Advances in Neural Information Processing Systems},
 editor = {A. Oh and T. Neumann and A. Globerson and K. Saenko and M. Hardt and S. Levine},
 pages = {9422--9457},
 publisher = {Curran Associates, Inc.},
 title = {White-Box Transformers via Sparse Rate Reduction},
 url = {https://proceedings.neurips.cc/paper_files/paper/2023/file/1e118ba9ee76c20df728b42a35fb4704-Paper-Conference.pdf},
 volume = {36},
 year = {2023}
}

@inproceedings{Viswarupan2024q,
  author={Viswarupan, Niruhan and Cheung, Gene and Lan, Fengbo and Brown, Michael S.},
  booktitle={IEEE International Conference on Acoustics, Speech and Signal Processing (ICASSP)},
  title={Mixed Graph Signal Analysis of Joint Image Denoising / Interpolation},
  year={2024},
  pages={9431-9435},
  doi={10.1109/ICASSP48485.2024.10445943}
}

@InProceedings{Liang_2021_ICCV,
    author    = {Liang, Jingyun and Cao, Jiezhang and Sun, Guolei and Zhang, Kai and Van Gool, Luc and Timofte, Radu},
    title     = {{SwinIR}: Image Restoration Using {Swin} Transformer},
    booktitle = {Proceedings of the IEEE/CVF International Conference on Computer Vision (ICCV) Workshops},
    month     = {October},
    year      = {2021},
    pages     = {1833-1844}
}

@inproceedings{zamir2022restormer,
  title={Restormer: Efficient Transformer for High-Resolution Image Restoration},
  author={Zamir, Syed Waqas and Arora, Aditya and Khan, Salman and Hayat, Munawar and Khan, Fahad Shahbaz and Yang, Ming-Hsuan},
  booktitle={Proceedings of the IEEE/CVF Conference on Computer Vision and Pattern Recognition},
  pages={5728--5739},
  year={2022}
}

@inproceedings{
do2024interpretable,
title={Interpretable Lightweight Transformer via Unrolling of Learned Graph Smoothness Priors},
author={VIET HO TAM THUC Do and Parham Eftekhar and Seyed Alireza Hosseini and Gene Cheung and Philip Chou},
booktitle={The Thirty-eighth Annual Conference on Neural Information Processing Systems},
volume={37},
pages={6393--6416},
year={2024},
url={https://openreview.net/forum?id=i8LoWBJf7j}
}

@inproceedings{agustsson2017ntire,
  title={NTIRE 2017 Challenge on Single Image Super-Resolution: Dataset and Study},
  author={Agustsson, Eirikur and Timofte, Radu},
  booktitle={Proceedings of the IEEE Conference on Computer Vision and Pattern Recognition (CVPR) Workshops},
  year={2017},
  pages={126--135}
}

@article{1992Bi,
  title={Bi-CGSTAB: A Fast and Smoothly Converging Variant of Bi-CG for the Solution of Nonsymmetric Linear Systems},
  author={ Vorst, Henk A. Van Der },
  journal={SIAM Journal on Scientific and Statistical Computing},
  volume={13},
  pages={631},
  year={1992},
}

@article{1952Methods,
  title={Methods of Conjugate Gradients for Solving Linear Systems},
  author={ Hestenes, Magnus R  and  Stiefel, E. L. },
  journal={Journal of Research of the National Bureau of Standards (United States)},
  volume={49},
  year={1952},
}

@inproceedings{bevilacqua2012low,
  title={Low-Complexity Color Image Reconstruction Based on A-Law Compressively Sampled Measurements},
  author={Bevilacqua, Marco and Roumy, Aline and Guillemot, Christine and Alberi-Morel, Marie-Line},
  booktitle={Proceedings of the British Machine Vision Conference (BMVC)},
  pages={25.1--25.10},
  year={2012},
  organization={BMVA Press},
  doi={10.5244/C.26.25}
}

@inproceedings{liu2020residual,
  title={Residual feature distillation network for lightweight image super-resolution},
  author={Liu, Jie and Tang, Jie and Wu, Gangshan},
  booktitle={Computer Vision--ECCV 2020: 16th European Conference},
  pages={41--55},
  year={2020},
  organization={Springer}
}

@inproceedings{kong2022residual,
  title={Residual local feature network for efficient super-resolution},
  author={Kong, Fangyuan and Li, Mingxi and Liu, Songwei and Liu, Ding and He, Jingwen and Bai, Yang and Chen, Fangya and Fu, Lean},
  booktitle={Proceedings of the IEEE/CVF Conference on Computer Vision and Pattern Recognition (CVPR) Workshops},
  pages={766--776},
  year={2022}
}

@inproceedings{sun2023spatially,
  title={Spatially-adaptive feature modulation for efficient image super-resolution},
  author={Sun, Long and Jiang, Jinshan and Wu, Jiang and He, Zhiqiang and Liu, Wentao and Zhang, Hongyan},
  booktitle={Proceedings of the IEEE/CVF International Conference on Computer Vision (ICCV)},
  pages={13190--13199},
  year={2023}
}
